\documentclass[conference]{IEEEtran}
\usepackage{times}

\usepackage[numbers,sort&compress]{natbib}
\usepackage{multicol}
\usepackage[bookmarks=true]{hyperref}
\usepackage{amsmath,amsfonts, amssymb, mathtools}
\usepackage{algorithmic}
\usepackage{algorithm}
\usepackage{array}
\usepackage{booktabs}
\usepackage[caption=false,font=normalsize,labelfont=sf,textfont=sf]{subfig}
\usepackage{textcomp}
\usepackage{textcase}
\usepackage{multirow}
\usepackage{stfloats}
\usepackage{balance}
\usepackage{orcidlink}
\usepackage{bm}
\usepackage{url}
\usepackage{verbatim}
\usepackage{graphicx}
\usepackage{stfloats}
\usepackage{booktabs}
\usepackage{multirow}
\usepackage{placeins}
\usepackage{color}
\newcommand{\best}[1]{\textbf{#1}}
\newcommand{\sbest}[1]{\underline{#1}}
\usepackage[font=small]{caption}
\begin{document}

\title{A Cross-Scale Decoder with Token Refinement for Off-Road Semantic Segmentation}

\author{
Seongkyu Choi \quad Jhonghyun An$^{*}$
}



%

\maketitle

\begin{abstract}
Off-road semantic segmentation is fundamentally challenged by irregular terrain, vegetation clutter, and inherent annotation ambiguity. Unlike urban scenes with crisp object boundaries, off-road environments exhibit strong class-level similarity among terrain categories, resulting in thick and uncertain transition regions that degrade boundary coherence and destabilize training. Rare or thin structures, such as narrow traversable gaps or isolated obstacles, further receive sparse and unreliable supervision and are easily overwhelmed by dominant background textures. Existing decoder designs either rely on low-scale bottlenecks that oversmooth fine structural details, or repeatedly fuse high-detail features, which tends to amplify annotation noise and incur substantial computational cost.
We present a cross-scale decoder that explicitly addresses these challenges through three complementary mechanisms. First, a global--local token refinement module consolidates semantic context on a compact bottleneck lattice, guided by boundary-aware regularization to remain robust under ambiguous supervision. Second, a gated detail bridge selectively injects fine-scale structural cues only once through cross-scale attention, preserving boundary and texture information while avoiding noise accumulation. Third, an uncertainty-guided class-aware point refinement selectively updates the least reliable pixels, improving rare and ambiguous structures with minimal computational overhead.
The resulting framework achieves noise-robust and boundary-preserving segmentation tailored to off-road environments, recovering fine structural details while maintaining deployment-friendly efficiency. Experimental results on standard off-road benchmarks demonstrate consistent improvements over prior approaches without resorting to heavy dense feature fusion.
\end{abstract}

\IEEEpeerreviewmaketitle

\section{Introduction}

\begin{figure}[t]
  \centering
  \includegraphics[width=\linewidth]{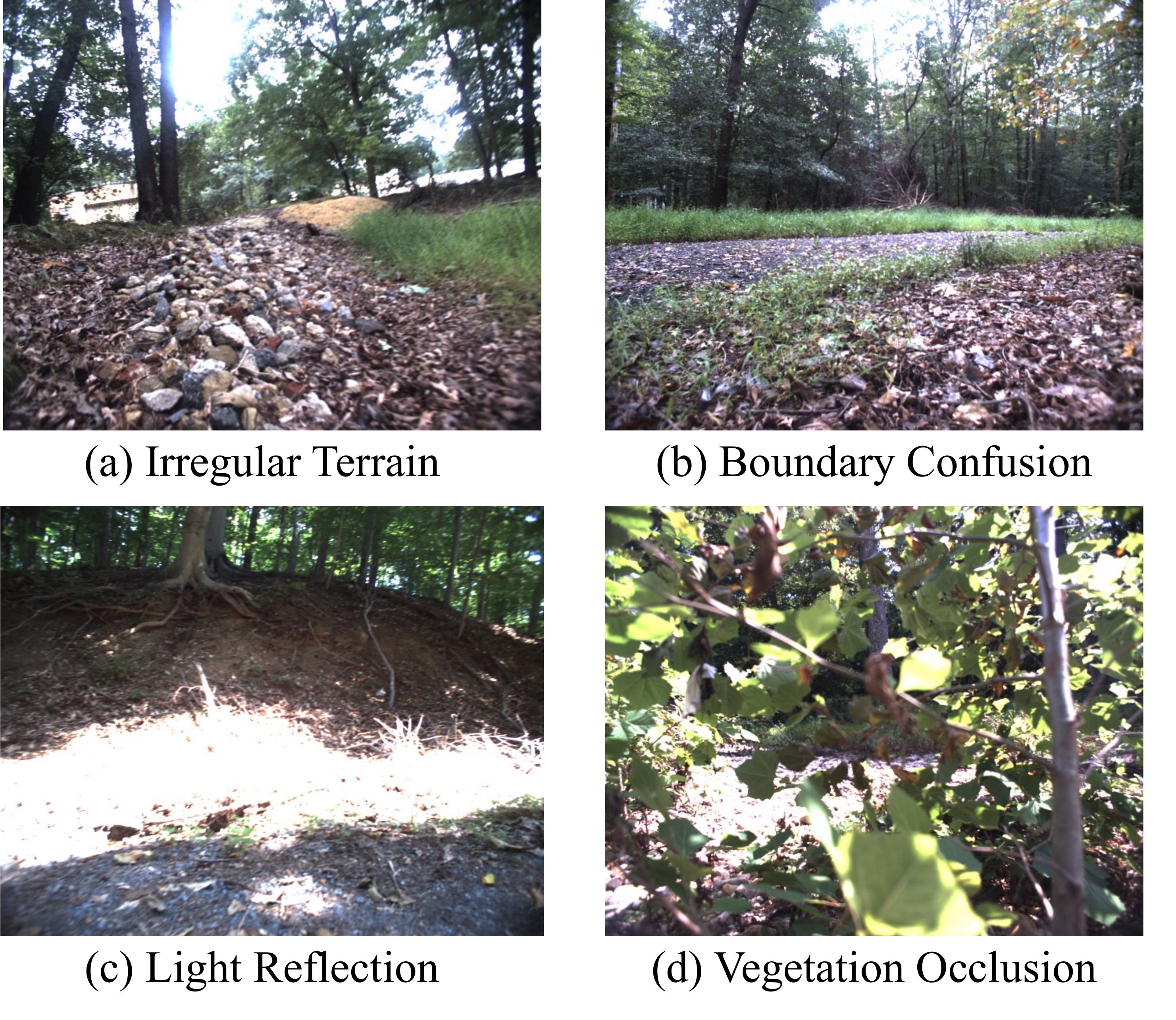}
  \caption{Off-road scenes exhibit diverse challenges, including irregular and uneven terrain, ambiguous boundaries between similar surfaces, strong light reflections, and severe occlusion by vegetation.}
  \label{fig:off_overview}
\end{figure}

Off-road semantic segmentation presents challenges that differ fundamentally from perception in structured urban environments. Irregular terrain, weak spatial organization, and inherently ambiguous semantic boundaries lead to annotations that are thick, inconsistent, and noisy, particularly around class transitions~\cite{wigness2019rugd, jiang2021rellis, mortimer2024goose, triest2022tartandrive, sivaprakasam2024tartandrive}. Vegetation clutter and appearance variations further exacerbate boundary ambiguity, making stable and boundary-coherent learning substantially more difficult than in urban scenes.

A key difficulty arises from strong class-level similarity. Many terrain categories share overlapping visual cues in color, texture, and scale, rendering both class identity and boundaries ambiguous in practice. Thin or rare structures, such as narrow traversable gaps or small obstacles, are often occluded and sparsely annotated, and even human annotators may disagree on precise boundaries, resulting in intrinsically weak and inconsistent supervision~\cite{guan2022ga, shaban2022semantic, gao2021fine}.

Robust off-road semantic segmentation is therefore critical for field robotics, where semantic perception directly informs downstream navigation and safety decisions under extreme class imbalance and sparse supervision~\cite{guan2022ga, shaban2022semantic, benatti2022end}. In such settings, errors near ambiguous boundaries or rare but safety-critical structures can propagate into unsafe traversal behavior, making robustness to label uncertainty a system-level requirement.

Despite recent advances, most existing segmentation architectures remain ill-suited to these conditions. Transformer-based and hybrid designs commonly aggregate multi-scale features into compact low-resolution bottlenecks, discarding fine-scale structural cues early in the network~\cite{zheng2021rethinking, xie2021segformer, ranftl2021vision}. While dense multi-scale fusion can partially recover lost details, it often amplifies annotation noise, increases computational cost, and introduces train--test mismatches when refinement branches are used only during training~\cite{zhao2017pyramid, chen2018encoder}. 
Recent foundation segmentation models, such as SAM-based approaches, demonstrate strong generalization for visually well-defined object boundaries; however, their dense mask generation and boundary sharpening assumptions are often misaligned with off-road scenes dominated by thick, ambiguous transitions and annotation noise~\cite{kirillov2023segment, zou2023segment, ravi2024sam}.

These limitations reveal a structural mismatch between existing decoder designs and the requirements of off-road perception. Robust segmentation under ambiguous supervision requires explicitly separating semantic consolidation from boundary-guided structural correction, rather than entangling both through repeated dense fusion~\cite{yuan2020object, takikawa2019gated, kirillov2020pointrend}.

To address this, we propose a Cross-Scale Decoder that decouples semantic consolidation from boundary-guided correction. Global--Local Token Refinement (GLTR) is first applied on a compact bottleneck lattice to stabilize global semantic representations under ambiguous supervision. Subsequently, boundary-guided correction is performed by selectively consulting fine-scale structural cues through a Gated Cross-Scale Interaction (GCS), which mediates semantic--structural information exchange without dense fusion. Within this interaction, uncertainty-guided, class-aware point-wise correction is applied to focus computation on a small set of low-confidence pixels after semantic stabilization and structural interaction. Together, these components preserve boundary geometry and thin structures while maintaining robustness and deployment efficiency in off-road robotic perception.

The main contributions of this work are summarized as follows:
\begin{itemize}
  \item We propose a Cross-Scale Decoder that explicitly separates semantic consolidation from boundary-guided correction, enabling robust off-road semantic segmentation under ambiguous and noisy supervision without relying on dense multi-scale fusion.
  \item We introduce Global--Local Token Refinement (GLTR), which stabilizes semantic representations on a compact bottleneck lattice through global attention and lightweight local refinement, supported by a boundary-band regularizer for noise-robust semantic consolidation.
  \item We realize boundary-guided correction via a Gated Cross-Scale Interaction (GCS), which selectively consults fine-scale structural cues and integrates uncertainty-guided, class-aware point-wise correction to recover rare and ambiguous structures with minimal computational overhead.
\end{itemize}

\section{Related Work}

\subsection{Off-Road Semantic Segmentation}

Semantic segmentation in off-road environments has gained increasing attention due to its importance in field robotics. Unlike urban datasets with well-defined object boundaries and structured layouts, off-road datasets exhibit irregular terrain, vegetation clutter, and significant annotation ambiguity. Class boundaries are often thick, inconsistent, and visually ambiguous, particularly around transitions between similar terrain types~\cite{wigness2019rugd, jiang2021rellis, guan2022ga, shaban2022semantic}.

This ambiguity reflects the intrinsic uncertainty of natural environments rather than simple annotation errors. Terrain categories frequently share overlapping visual characteristics, and fine-scale structures are often occluded or sparsely represented. As a result, dense and fully consistent pixel-level ground truth is difficult to obtain in practice, posing fundamental challenges to learning stable and boundary-coherent representations in off-road scenes~\cite{guan2022ga, gao2021fine, singh2021offroadtranseg}.

\subsection{Multi-Scale Decoder Architectures}

Recent advances in semantic segmentation have been driven by transformer-based and hybrid architectures that aggregate multi-scale features through attention mechanisms~\cite{zheng2021rethinking, xie2021segformer, liu2021swin, jain2023oneformer, cheng2022masked}. For computational efficiency, many designs compress feature representations into compact low-resolution bottlenecks before global reasoning. While effective for capturing long-range context, this strategy discards fine-scale structural cues early in the network~\cite{zheng2021rethinking, ranftl2021vision, xie2021segformer}.

Once thin structures and local discontinuities are suppressed at the bottleneck stage, they are difficult to recover through downstream refinement. Several approaches attempt to address this limitation through dense multi-scale fusion or skip connections~\cite{zhao2017pyramid, chen2018encoder, yuan2020object, fu2019dual}. However, repeated fusion often amplifies annotation noise and significantly increases computational cost, limiting suitability for resource-constrained robotic deployment~\cite{wu2020cgnet, poudel2019fast, yu2021bisenet}.

\subsection{Boundary-Aware Refinement}

Prior work has explored boundary-aware losses, affinity-based constraints, and post-processing techniques such as conditional random fields to improve boundary quality~\cite{krahenbuhl2011efficient, kervadec2019boundary, ke2018adaptive, wang2022active}. While effective at sharpening predictions, these methods typically operate as auxiliary components and do not address the architectural causes of boundary degradation.

More recent approaches adopt sparse or point-based refinement strategies to selectively correct uncertain regions~\cite{kirillov2020pointrend, yuan2020segfix, ding2019boundary}. However, many such modules are enabled only during training and disabled at inference to reduce latency, introducing train--test mismatches that can lead to unstable boundary behavior under noisy supervision~\cite{takikawa2019gated}.

Taken together, prior work suggests that improving off-road segmentation does not primarily require increasing model capacity or repeated dense fusion, but rather rethinking how semantic consolidation and structural correction are integrated within the decoder. In environments with intrinsic class-level ambiguity, a robust solution should preserve fine-scale boundary cues through selective, gated interaction while remaining fully active at inference time, motivating the Cross-Scale Decoder proposed in this work.

\begin{figure*}[t]
  \centering
  \includegraphics[width=\textwidth]{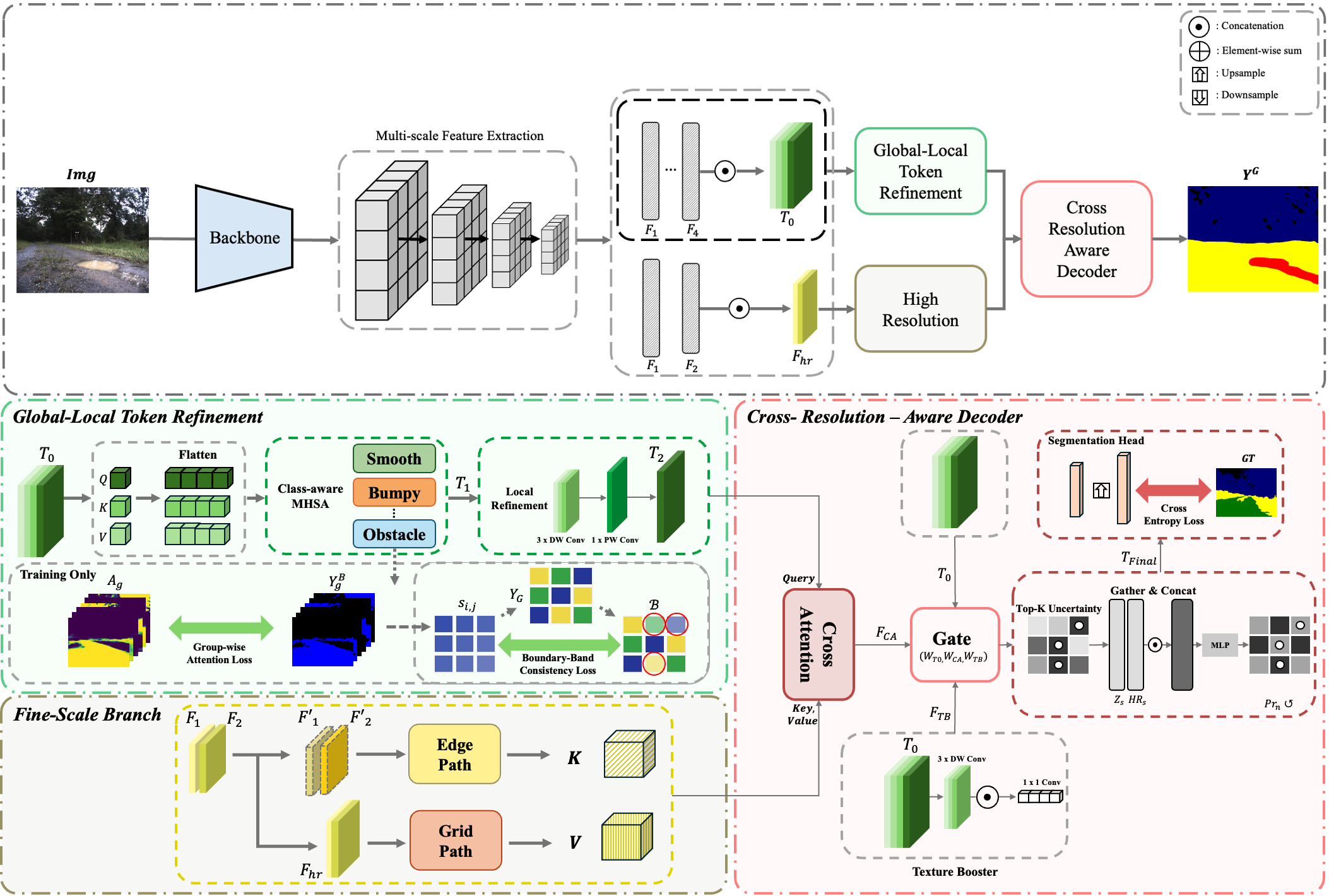}
\caption{Overall framework of our CSTR. Multi-scale backbone features are compressed into a compact bottleneck token $T_0$ and refined by Global--Local Token Refinement (GLTR). Boundary-Guided Correction extracts fine-scale structural cues, which are selectively consulted via a Gated Cross-Scale Interaction (GCS) to recover thin or ambiguous structures efficiently.}
  \label{fig:overview}
\end{figure*}

\section{Methodology}

\subsection{Overview}
Given an input image $I$, our method extracts multi-scale features using a transformer backbone and processes them with a Cross-Scale Decoder. Unlike conventional decoders that rely on repeated dense fusion, the proposed design explicitly constrains when and where structural information is introduced, reflecting the intrinsic ambiguity and noisy supervision of off-road environments. The decoder consolidates global semantics at a compact bottleneck while selectively consulting structural cues only when necessary, thereby explicitly separating semantic consolidation from boundary-guided structural correction.
As shown in Fig.~\ref{fig:overview}, the decoder consists of three conceptual components: Global--Local Token Refinement (GLTR) for stabilizing semantic representations, Boundary-Guided Correction (BGC) for extracting boundary-relevant structural cues, and a Gated Cross-Scale Interaction (GCS) that selectively integrates these cues and performs sparse point-wise correction.

\subsection{Global--Local Token Refinement (GLTR)}
Global--Local Token Refinement (GLTR), shown in the green block of Fig.~\ref{fig:overview}, stabilizes semantic representations before any structural correction is applied. In off-road scenes, premature incorporation of fine-scale cues can entangle noisy boundary information with global semantics. GLTR addresses this issue by consolidating class-aware context at a compact bottleneck lattice.

Multi-scale backbone features $\{F_l\}_{l=1}^{L}$ are first projected into a shared embedding space and aggregated into a compact bottleneck representation:

\begin{equation}
\label{eq:bottleneck}
T_0 = \sum_{l=1}^{L} \alpha_l \cdot \psi_l(F_l),
\end{equation}
where $\psi_l(\cdot)$ denotes a scale-specific projection and $\alpha_l$ controls each scale’s contribution. As indicated by Eq.~(\ref{eq:bottleneck}), semantic aggregation is intentionally performed before introducing any boundary-related cues, preventing early contamination of global semantics under ambiguous supervision.

\begin{equation}
\label{eq:bottleneck}
\begin{aligned}
      { T_0 = \sum_{l=1}^{L} \text{Softmax}(\mathbf{w}_{\alpha}^\top \mathcal{G}(F_l)) \cdot \phi_l(F_l) } 
\end{aligned}
\end{equation}
 
Here, $\mathcal{G}(\cdot)$ denotes the global average pooling operation, and $\mathbf{w}_{\alpha}$ represents a learnable projection vector that dynamically determines the importance of each scale. The function $\phi_l(\cdot)$ projects the scale-specific features $F_l$ into a unified embedding space $\mathbb{R}^{C \times H_0 \times W_0}$ via convolution and interpolation. This formulation allows the bottleneck $T_0$ to selectively aggregate multi-scale contexts based on their semantic relevance.

GLTR then performs class-aware semantic consolidation using scaled dot-product attention~\cite{vaswani2017attention}. For each class $c$, a class-specific semantic token is computed as
\begin{equation}
\label{eq:class_token}
T_0^{(c)} = \sum_{i=1}^{N} \alpha_i^{(c)} \, T_0^{(i)} .
\end{equation}

\begin{equation}
\label{eq:class_attention}
{ T_1 = \text{Softmax}\left(\frac{Q_c K^\top}{\sqrt{d_k}}\right) V,}
\end{equation}
 
We formulate class-aware consolidation as a cross-attention operation. Let $Q_c \in \mathbb{R}^{N_{class} \times d}$ be the set of learnable class prototypes. The keys $K$ and values $V$ are projected from the bottleneck token $T_0$ using linear transformations $W_k$ and $W_v$, respectively. The term $\sqrt{d_k}$ serves as a scaling factor to stabilize gradients. This mechanism explicitly queries the spatial feature map for class-specific semantic cues, effectively filtering out irrelevant background noise.

The refined semantic lattice is obtained by aggregating class-aware tokens and restoring local coherence through a lightweight refinement operation:
\begin{equation}
\label{eq:gltr}
T_2 = T_1 + \varphi(T_1),
\qquad
T_1 = \sum_{c=1}^{C} T_0^{(c)},
\end{equation}
where $\varphi(\cdot)$ denotes a depthwise--pointwise convolution block. As summarized in Eq.~(\ref{eq:gltr}), GLTR establishes a stable semantic core independently of fine-scale structural cues, ensuring that subsequent structural correction operates on already consolidated semantics rather than noisy intermediate representations.

During training, a boundary-band regularizer constrains class-aware attention updates to narrow transition regions, stabilizing token alignment near ambiguous boundaries while leaving inference unchanged. The refined tokens $T_2$ thus serve as the semantic foundation for subsequent boundary-guided correction.

\begin{figure}[t]
  \centering
  \includegraphics[width=0.75\linewidth]{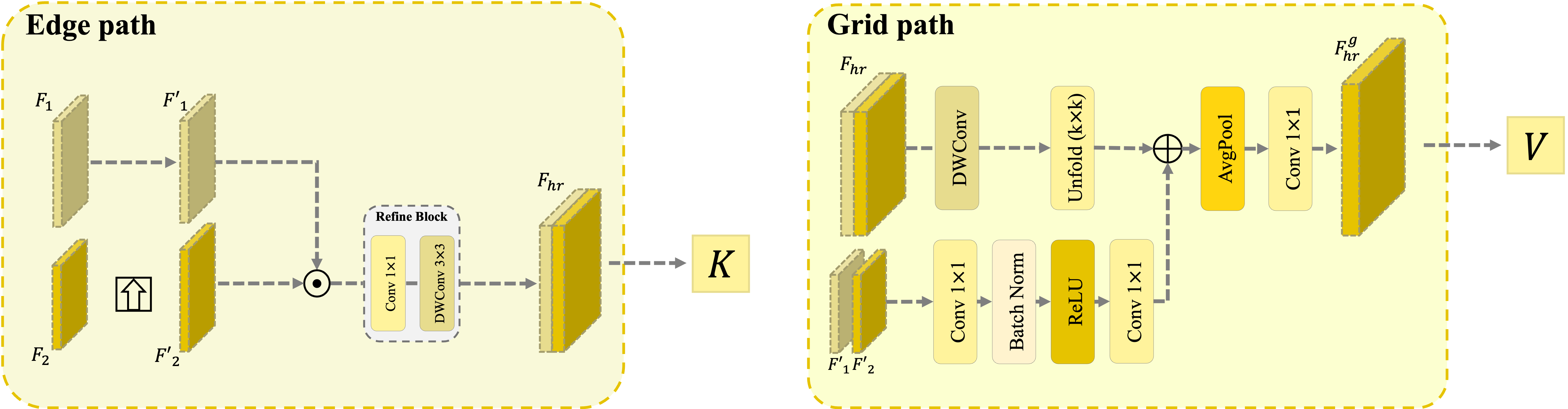}
  \caption{Structure of the Boundary-Guided Correction module. Boundary-sensitive cues are extracted through an edge path, while smoothed contextual cues are obtained via a grid path. These fine-scale structural cues are later consulted through gated cross-scale interaction without dense fusion.}
  \label{fig:fs_module}
\end{figure}

\subsection{Boundary-Guided Correction (BGC)}
Boundary-Guided Correction (BGC), shown in the yellow block of Fig.~\ref{fig:overview}, extracts fine-scale structural cues that are informative near ambiguous boundaries and thin structures. Rather than reinjecting these cues directly into the semantic stream, BGC isolates boundary-relevant information and defers its utilization to a later interaction stage.

As illustrated in Fig.~\ref{fig:fs_module}, BGC employs two complementary paths operating on early-stage features $F_s$: an edge path emphasizing boundary-sensitive structures and a grid path capturing smoothed contextual information. These paths produce a buffered set of boundary-guided structural cues:
\begin{equation}
\label{eq:bgc}
\mathcal{S} = \{K_s, V_s\}
=
\left\{
f_{\text{edge}}(F_s),\;
f_{\text{grid}}(F_s)
\right\},
\end{equation}
where $f_{\text{edge}}(\cdot)$ and $f_{\text{grid}}(\cdot)$ denote edge-preserving and grid-smoothing operations. As defined in Eq.~(\ref{eq:bgc}), $\mathcal{S}$ is retained as a buffer rather than being directly fused with semantic representations, preventing early contamination of global semantics with noisy boundary details.

\subsection{Gated Cross-Scale Interaction (GCS)}
The Gated Cross-Scale Interaction (GCS), shown in the red block of Fig.~\ref{fig:overview}, selectively integrates boundary-guided structural cues with stabilized semantic representations obtained from GLTR. Instead of dense fusion, semantic tokens query the buffered structural cues through cross-scale attention:
\begin{equation}
\label{eq:crossattn}
F_{\mathrm{cs}} = \mathrm{Attn}(Q=T_2,\; K=K_s,\; V=V_s),
\end{equation}
allowing structural information to be consulted only when relevant and limiting unnecessary noise propagation.

The queried structural information is then selectively injected into the semantic stream via a gated formulation:
\begin{equation}
\label{eq:gcs}
T_3 = T_2 + g(T_2, \mathcal{S}) \odot F_{\mathrm{cs}},
\end{equation}
where the gating function is defined as
\begin{equation}
g(T_2, \mathcal{S}) =
\sigma\!\left(W_T T_2 + W_S F_{\mathrm{cs}}\right).
\end{equation}
As formalized in Eq.~(\ref{eq:gcs}), boundary-guided structural cues influence semantic representations only when beneficial, enabling single-shot structural correction while preventing excessive amplification of boundary noise. This selective design avoids the instability commonly caused by repeated structural injection under ambiguous supervision.

Within this interaction, uncertainty-guided, class-aware point-wise correction is further integrated to address residual ambiguities that cannot be resolved through token-level interaction alone. Only a small set of pixels with high prediction uncertainty is selected, and their predictions are refined using a lightweight MLP conditioned on both semantic features from $T_3$ and corresponding structural cues. By restricting refinement to uncertain regions, the decoder improves boundary precision and thin structures with minimal additional computational overhead.

\begin{table*}[!t]
  \centering
  \small
  \setlength{\tabcolsep}{5.2pt}
  \renewcommand{\arraystretch}{1.1}
  \caption{Comparison on \textsc{RUGD} and \textsc{RELLIS-3D}. We report per-group IoU (\%), mean IoU (\(\mathrm{mIoU}\!\uparrow\)), and average accuracy (\(\mathrm{aAcc}\!\uparrow\)). Asterisks (*) denote transformer-based methods. Best per dataset in \textbf{bold}, second best \underline{underlined}.}
  \label{tab:sota}
  \begin{tabular}{@{}cl*{8}{c}@{}}
    \toprule
    \textbf{Dataset} & \textbf{Methods (IoU)} &
    \textbf{Smooth} & \textbf{Rough} & \textbf{Bumpy} &
    \textbf{Forbidden} & \textbf{Obstacle} & \textbf{Background} &
    \textbf{mIoU$\uparrow$} & \textbf{aAcc$\uparrow$} \\
    \midrule
    \multirow[c]{15}{*}{\textbf{RUGD}}
    & PSPNet~\cite{zhao2017pyramid}         & 48.62 & 88.92 & 69.45 & 29.07 & 87.98 & 78.29 & 67.06 & 92.85 \\
    & DeepLabv3+~\cite{chen2018encoder}     &  5.86 & 84.99 & 50.40 & 25.04 & 87.50 & \best{81.47} & 55.88 & 91.51 \\
    & DANet~\cite{fu2019dual}               &  2.26 & 81.47 &  8.69 & 15.00 & 82.54 & 74.86 & 44.14 & 88.81 \\
    & OCRNet~\cite{yuan2020object}    & 66.29 & 89.47 & 76.15 & 59.14 & 88.77 & 79.17 & 76.50 & 93.46 \\
    & PSANet~\cite{zhao2018psanet}          & 34.92 & 87.70 & 35.64 &  8.66 & 86.95 & 78.97 & 55.47 & 92.13 \\
    & BiSeNetv2~\cite{yu2021bisenet}        & 24.27 & 89.99 & \sbest{89.99} & 83.31 & 90.93 & 75.29 & 75.10 & 93.40 \\
    & CGNet~\cite{wu2020cgnet}              & 40.84 & 90.39 & 85.67 & 76.21 & 89.75 & 74.48 & 76.22 & 93.29 \\
    & FastSCNN~\cite{poudel2019fast}        & 83.03 & 92.82 & 87.69 & 81.05 & 90.94 & 75.11 & 85.11 & 94.77 \\
    & FastFCN~\cite{wu2019fastfcn}          & 26.27 & 89.85 & 85.95 & 84.13 & 91.23 & 75.63 & 75.51 & 93.46 \\
    & *SETR~\cite{zheng2021rethinking}      & 89.77 & 92.46 & 84.58 & 70.33 & 89.55 & 70.47 & 82.86 & 94.09 \\
    & *DPT~\cite{ranftl2021vision}          &  1.04 & 81.23 & 22.98 & 25.84 & 89.18 & 74.50 & 49.13 & 88.77 \\
    & *SegFormer~\cite{xie2021segformer}    & 93.26 & 93.16 & 87.56 & 77.31 & 91.20 & 78.50 & 86.83 & 95.17 \\
    & *SegNeXt~\cite{guo2022segnext}        & 90.39 & 91.17 & 83.96 & 65.43 & 87.80 & 68.17 & 81.15 & 93.22 \\
    & *GA-Nav~\cite{guan2022ga}             & \best{95.15} & \sbest{94.45} & 89.83 & \sbest{86.25} & \sbest{91.95} & 76.86 & \sbest{89.08} & \sbest{95.66} \\
    \cmidrule(lr){2-10}
    & \textbf{*CSTR (ours)}                 & \sbest{95.11} & \best{94.47} & \best{90.36} & \best{87.16} & \best{92.60} & \sbest{80.13} & \best{89.97} & \best{95.98} \\
    \midrule
    \multirow[c]{15}{*}{\textbf{RELLIS-3D}}
    & PSPNet~\cite{zhao2017pyramid}         & 69.21 & 80.99 &  8.89 & 53.70 & 60.70 & 94.67 & 61.36 & 86.01 \\
    & DeepLabv3+~\cite{chen2018encoder}     & 65.76 & 79.84 & 19.72 & 47.52 & 64.88 & 95.92 & 62.27 & 85.84 \\
    & DANet~\cite{fu2019dual}               & 72.93 & 85.18 & 13.10 & 60.60 & 70.53 & 95.65 & 66.38 & 89.11 \\
    & OCRNet~\cite{yuan2020object}    & 74.67 & 83.04 & 27.76 & 60.44 & 62.35 & 92.58 & 66.81 & 86.95 \\
    & PSANet~\cite{zhao2018psanet}          & 64.06 & 75.29 & 17.08 & 47.45 & 61.74 & 94.31 & 59.99 & 83.71 \\
    & BiSeNetv2~\cite{yu2021bisenet}        & 65.56 & 73.24 & 39.35 & 48.17 & 71.91 & 93.78 & 65.33 & 83.03 \\
    & CGNet~\cite{wu2020cgnet}              & 62.84 & 74.17 & 49.57 & 45.41 & 68.88 & 94.53 & 65.90 & 82.70 \\
    & FastSCNN~\cite{poudel2019fast}        & 67.06 & 77.60 & \best{56.49} & 49.76 & 70.31 & 94.43 & 69.27 & 84.51 \\
    & FastFCN~\cite{wu2019fastfcn}          & 70.51 & 79.15 & 49.72 & 51.37 & 63.90 & 94.82 & 68.24 & 84.10 \\
    & *SETR~\cite{zheng2021rethinking}      & 65.37 & 78.64 & 40.89 & 52.59 & 63.80 & 91.87 & 65.53 & 83.59 \\
    & *DPT~\cite{ranftl2021vision}          &  5.42 & 76.65 & 47.13 & 54.87 & 62.74 & 85.50 & 55.38 & 81.61 \\
    & *SegFormer~\cite{xie2021segformer}    & 60.28 & 79.78 & 53.35 & 53.78 & 70.15 & 94.37 & 68.62 & 85.37 \\
    & *SegNeXt~\cite{guo2022segnext}        & 51.67 & 78.40 & 19.38 & 42.61 & 66.04 & 92.05 & 58.36 & 82.16 \\
    & *GA-Nav~\cite{guan2022ga}             & \sbest{78.50} & \sbest{88.25} & \sbest{37.28} & \sbest{72.34} & \sbest{74.75} & \sbest{96.07} & \sbest{74.44} & \sbest{91.69} \\
    \cmidrule(lr){2-10}
    & \textbf{*CSTR (ours)}                 & \best{80.92} & \best{89.16} & 37.26 & \best{73.56} & \best{75.08} & \best{96.35} & \best{75.39} & \best{92.15} \\
    \bottomrule
  \end{tabular}
\end{table*}

\begin{table}[t]
  \centering
  \caption{\protect\NoCaseChange{
  Unified grouping of fine-grained RUGD and RELLIS-3D labels into a 6-class terrain hierarchy
  based on surface texture, navigability, and semantic characteristics.}}
  \label{tab:terrain-groups-unified}
  \renewcommand{\arraystretch}{1.05}
  \begin{tabular}{@{}p{.22\columnwidth} p{.36\columnwidth} p{.36\columnwidth}@{}}
    \toprule
    \textbf{Terrain Group} 
      & \textbf{RUGD Labels} 
      & \textbf{RELLIS-3D Labels} \\
    \midrule
    Smooth   
      & Concrete, asphalt 
      & Concrete, asphalt \\

    Rough     
      & Gravel, grass, dirt, sand 
      & Dirt, grass \\

    Bumpy     
      & Rock, rock bed 
      & Mud, rubble \\

    Forbidden 
      & \mbox{Water, bushes, tall vegetation} 
      & Water, bush \\

    Obstacles        
      & Trees, poles, logs, etc. 
      & Tree, pole, vehicle, etc. \\

    Background       
      & Void, sky, signs 
      & Void, sky \\
    \bottomrule
  \end{tabular}
  \vspace{-6pt}
\end{table}

\section{Experiments}

\subsection{Datasets}

We evaluate our method on two off-road semantic segmentation benchmarks, RUGD~\cite{wigness2019rugd} and RELLIS-3D~\cite{jiang2021rellis}.
Both datasets capture unstructured outdoor environments with heterogeneous terrain, ambiguous boundaries, and strong class imbalance, making them suitable for evaluating robustness under noisy supervision~\cite{wigness2019rugd, jiang2021rellis, guan2022ga}.

\textbf{RUGD.}
This dataset contains diverse off-road RGB scenes with irregular terrain geometry and abundant thin structures.
Annotations often exhibit thick or ambiguous transitions between terrain classes, posing challenges for dense high-resolution fusion.
We follow the official train/test split and use RGB inputs only.
For navigation-oriented evaluation, fine-grained labels are merged into six terrain groups:
Smooth, Rough, Bumpy, Forbidden, Obstacle, and Background (Table~\ref{tab:terrain-groups-unified}).

\textbf{RELLIS-3D.}
This dataset is collected using an unmanned ground vehicle and provides longer sequences with recurring terrain patterns.
Despite being slightly more structured than RUGD, it still exhibits substantial label noise and boundary ambiguity due to vegetation occlusion and visually similar surfaces.
We adopt the standard split and map all categories to the same six-class hierarchy used for RUGD (Table~\ref{tab:terrain-groups-unified}) to ensure consistent evaluation.

For both datasets, images are resized to match the decoder input resolution (300$\times$375 for RUGD and 375$\times$600 for RELLIS-3D).
Standard data augmentation (random cropping, horizontal flipping, and color jitter) is applied during training, while center cropping followed by resizing is used at test time.

\begin{figure}[t]
  \centering
  \includegraphics[width=\linewidth]{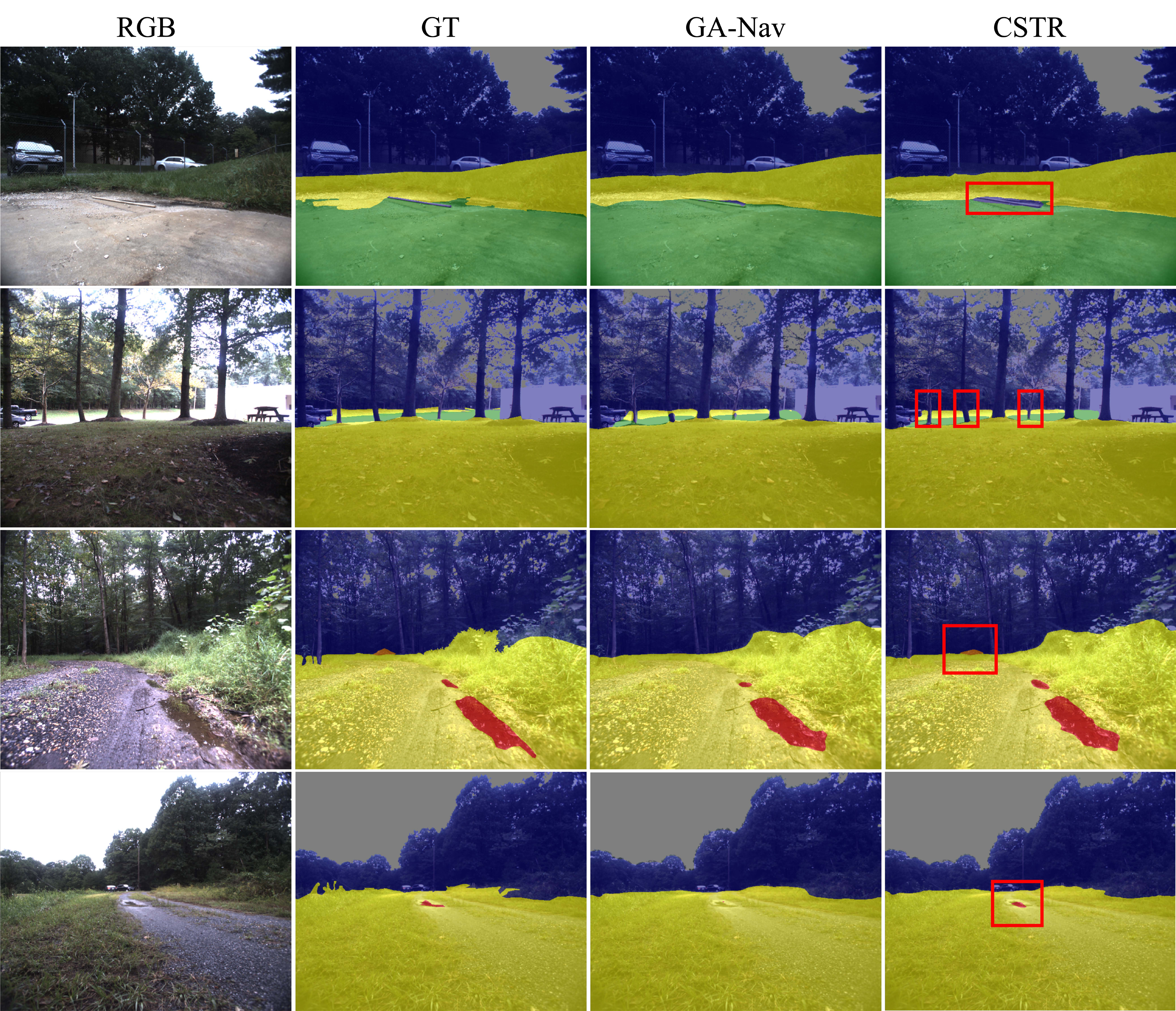}
  \caption{Qualitative comparison on RUGD highlighting rare and thin structures.
Compared to GA-Nav, CSTR better preserves small objects and narrow regions while maintaining coherent terrain boundaries.}
  \label{fig:qual_rugd}
\end{figure}

\subsection{Experiment Settings}

All experiments are implemented in PyTorch using the MMSegmentation framework~\cite{contributors2020mmsegmentation}
and conducted on a single NVIDIA RTX A5000 GPU with mixed-precision training.
We adopt MiT-B0~\cite{xie2021segformer} as the backbone, initialized with ImageNet pre-trained weights~\cite{deng2009imagenet},
and attach our Cross-Scale Decoder to the same backbone configuration used by GA-Nav~\cite{guan2022ga} for fair comparison.

Models are trained for 240k iterations using SGD with momentum 0.9 and weight decay $4\times10^{-5}$.
A polynomial learning rate schedule (power 0.9) with linear warm-up over the first 1.5k iterations is employed.
The initial learning rate is set to $6\times10^{-2}$ for RUGD and $3\times10^{-3}$ for RELLIS-3D.
Batch sizes are 8 for RUGD and 2 for RELLIS-3D under the same hardware configuration.
Synchronized Batch Normalization and gradient clipping (max norm 35) are applied for training stability.

\subsection{Main Results}

We compare the proposed Cross-Scale Decoder (CSTR) with representative CNN- and Transformer-based baselines commonly used for semantic segmentation~\cite{zhao2017pyramid, chen2018encoder, xie2021segformer, guan2022ga} on RUGD~\cite{wigness2019rugd} and RELLIS-3D~\cite{jiang2021rellis}.
Quantitative results are summarized in Table~\ref{tab:sota}.
Across both datasets, CSTR achieves the highest mIoU and aAcc, demonstrating consistent improvements over prior methods.

On RUGD, CSTR improves upon GA-Nav~\cite{guan2022ga} in both mIoU and aAcc, with notable gains on visually ambiguous classes such as \textit{Forbidden} and \textit{Background}.
These classes are prone to boundary leakage around vegetation and reflective regions, indicating that selective fine-scale consultation and boundary-aware supervision effectively reduce over-smoothing.
These improvements are consistent with our design choice of consolidating semantics at a compact bottleneck and applying single-shot, gated fine-scale correction under ambiguous supervision.

On RELLIS-3D, CSTR also outperforms GA-Nav, with gains concentrated on terrain groups with irregular geometry and sparse annotations, including \textit{Forbidden} and \textit{Obstacle}.
This suggests that avoiding repeated structural fusion and instead relying on a single gated interaction is particularly effective under severe annotation ambiguity.

\begin{figure}[!t]
  \centering
  \includegraphics[width=\linewidth]{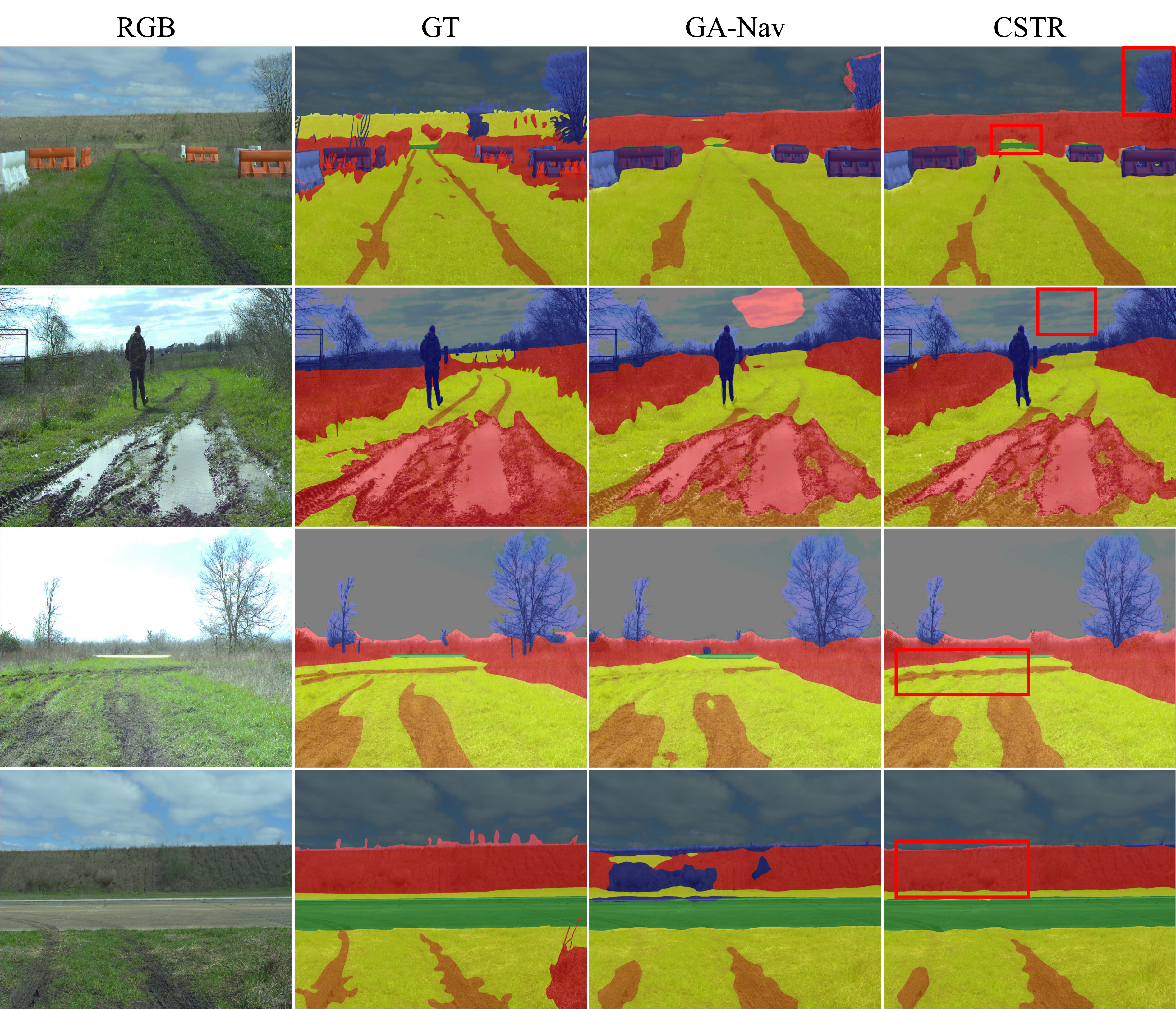}
  \caption{Qualitative comparison on RELLIS-3D under visually ambiguous terrain transitions.
  CSTR produces more coherent region interiors and smoother class boundaries around obstacles and vegetation compared to GA-Nav.}
  \label{fig:qual_rellis}
\end{figure}

Qualitative examples in Fig.~\ref{fig:qual_rugd} and Fig.~\ref{fig:qual_rellis} corroborate these results.
Compared to GA-Nav and Transformer-based decoders, CSTR produces sharper transitions, cleaner region interiors, and fewer boundary artifacts, while preserving interior semantic consistency.

Overall, these results demonstrate that consolidating global semantics at a compact bottleneck and selectively correcting fine-scale ambiguities yields robust performance gains for off-road semantic segmentation under noisy supervision.

\begin{table}[!t]
\centering
\caption{Ablation of gated fusion configurations on RUGD, analyzing the effect of different gating inputs on semantic accuracy and boundary-sensitive performance.}
\label{tab:gate_ablation}
\setlength{\tabcolsep}{8pt}      
\renewcommand{\arraystretch}{1.3} 
\begin{tabular}{llcccc}
\hline
Gate & Inputs to Gate & mIoU$\uparrow$ & bIoU$\uparrow$ & F1$\uparrow$ & aAcc$\uparrow$ \\
\hline
\multirow{1}{*}{1-way} & CA          & 89.72 & 32.22 & 48.87 & 95.93 \\
\multirow{2}{*}{2-way} & CA+$T_0$       & 89.71 & 32.12 & 48.78 & 95.83 \\
                       & CA+TB       & 89.86 & 32.62 & 49.14 & 95.95 \\
\multirow{1}{*}{3-way} & CA+TB+$T_0$    & 89.97 & 32.75 & 49.18 & 95.98 \\
\hline
\end{tabular}
\end{table}

\begin{table}[!t]
\centering
\caption{Incremental ablation results on RUGD demonstrating the contribution of each decoder component. 
We report region-level metrics ($\mathrm{mIoU}$, $\mathrm{aAcc}$) together with boundary- and structure-sensitive metrics ($\mathrm{bIoU}$, $F_1$).}
\label{tab:ablation}
\setlength{\tabcolsep}{6pt}
\renewcommand{\arraystretch}{1.3} 
\begin{tabular}{lcccc}
\hline
Variant & mIoU$\uparrow$ & bIoU$\uparrow$ & F1$\uparrow$ & aAcc$\uparrow$ \\
\hline
Baseline                   & 88.32 & 28.53 & 46.97 & 95.19 \\
\,+ GLTR                   & 88.66 & 29.78 & 47.81 & 95.44 \\
\,+ BGC                    & 88.80 & 29.75 & 47.80 & 95.51 \\
\,+ GCS (w/o point-wise)   & 88.86 & 30.03 & 48.05 & 95.52 \\
\,+ GCS (w/ point-wise)    & 89.97 & 32.75 & 49.18 & 95.98 \\
\hline
\end{tabular}
\end{table}

\subsection{Qualitative Assessment}
Qualitative results demonstrate that the proposed CSTR produces more coherent and stable segmentations than existing decoders across both RUGD~\cite{wigness2019rugd} and RELLIS-3D~\cite{jiang2021rellis}, particularly in challenging scenarios with ambiguous boundaries, cluttered backgrounds, and sparse supervision.
These qualitative improvements reflect the effect of GLTR in stabilizing global semantics and BGC/GCS in selectively correcting boundary regions without repeated dense fusion.

On RUGD, examples in Fig.~\ref{fig:qual_rugd} highlight CSTR’s ability to preserve rare and thin structures, such as small objects and narrow traversable regions, while maintaining continuous terrain boundaries in cluttered scenes.
Compared to GA-Nav~\cite{guan2022ga} and Transformer-based baselines, CSTR reduces fragmentation and boundary leakage between visually similar classes (e.g., soil, gravel, and grass), reflecting the benefit of selectively consulting fine-scale structural cues rather than repeatedly aggregating detailed features.

On RELLIS-3D, Fig.~\ref{fig:qual_rellis} demonstrates that CSTR generalizes effectively to environments with man-made obstacles, irregular terrain geometry, and partial occlusions.
CSTR produces cleaner region interiors and more structurally consistent predictions around obstacles and vegetation, even when class boundaries are visually ambiguous or sparsely annotated.
This cross-dataset robustness indicates that single-shot fine-scale consultation and class-aware sparse refinement remain effective beyond the training domain.

Overall, these qualitative observations are consistent with the quantitative improvements reported in Table~\ref{tab:sota}, confirming that compact semantic consolidation followed by selective fine-scale correction yields reliable and noise-robust off-road semantic segmentation.

\subsection{Ablation Studies}

\noindent\textbf{Incremental Module Analysis.}
We conduct incremental ablations on the RUGD dataset to evaluate the contribution of each decoder component, progressively enabling modules following the architectural design order.

As reported in Table~\ref{tab:ablation}, each component contributes consistent improvements across both region-level metrics (\(\mathrm{mIoU}\), \(\mathrm{aAcc}\)) and boundary- and structure-sensitive metrics (\(\mathrm{bIoU}\), \(F_1\)).
Here, bIoU evaluates prediction consistency within a narrow band around class boundaries~\cite{cheng2021boundary},
while the reported \(F_1\) score emphasizes rare and thin structures, capturing fragmentation behavior not reflected by boundary-based metrics.

The results highlight the complementary roles of the proposed components.
GLTR stabilizes semantic representations at a compact bottleneck under ambiguous supervision, leading to consistent gains in region-level accuracy.
Introducing BGC further improves boundary localization by extracting boundary-relevant structural cues without densely reinjecting them into the semantic stream.
Subsequent GCS enables selective semantic--structural integration, yielding additional gains in boundary coherence.
Finally, incorporating point-wise correction within GCS amplifies these improvements by explicitly refining uncertain predictions, resulting in pronounced gains on rare and thin structures as reflected in the \(F_1\) score.

The gradual increase in bIoU across modules indicates that boundary stability benefits from progressive semantic consolidation followed by selective structural correction.
Overall, these trends validate that the proposed decoder components address complementary aspects of off-road semantic segmentation, including semantic ambiguity, boundary uncertainty, and structural sparsity.

\begin{figure}[t]
  \centering
  \includegraphics[width=\linewidth]{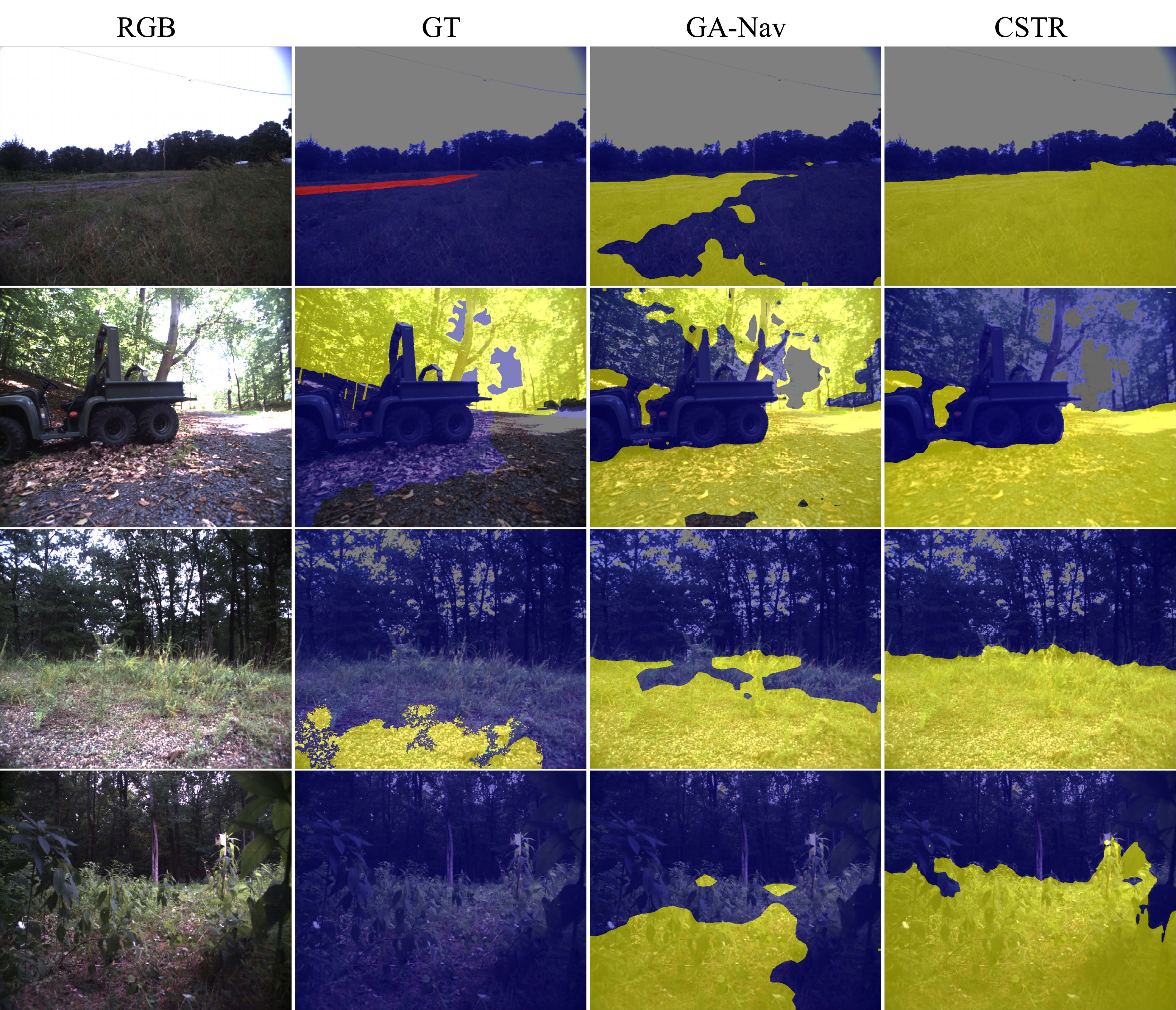}
  \caption{Quantitative robustness evaluation on RUGD with noisy ground truth.
CSTR exhibits smaller performance degradation than GA-Nav under increasing label noise.}
  \label{fig:qual_gtnoise}
\end{figure}

\begin{table}[t]
\centering
\caption{Qualitative comparison on RUGD with noisy ground truth. CSTR produces more coherent and structurally consistent predictions than GA-Nav under label noise.}
\label{tab:noise_robustness_compare}
\setlength{\tabcolsep}{6pt}
\renewcommand{\arraystretch}{1.3}
\begin{tabular}{llcccc}
\hline
Model & Noise Level & mIoU$\uparrow$ & bIoU$\uparrow$ & $F1_{\text{boundary}}\uparrow$ & aAcc$\uparrow$ \\
\hline
\multirow{4}{*}{GA-Nav} 
& clean & 89.08 & 39.66 & 93.65 & 95.66 \\
& $r=1$ & 87.66 & 39.34 & 93.28 & 95.20 \\
& $r=3$ & 88.36 & 39.24 & 93.69 & 95.43 \\
& $r=5$ & 88.05 & 39.84 & 93.51 & 95.60 \\
\hline
\multirow{4}{*}{Ours} 
& clean & 89.97 & 45.94 & 94.64 & 95.98 \\
& $r=1$ & 89.18 & 43.55 & 94.19 & 95.73 \\
& $r=3$ & 88.85 & 43.29 & 94.01 & 95.62 \\
& $r=5$ & 88.20 & 42.47 & 93.62 & 95.30 \\
\hline
\end{tabular}
\end{table}

\vspace{2pt}
\noindent\textbf{Gated Interaction Configuration.}
We further analyze different gating configurations within the Gated Cross-Scale Interaction module (Table~\ref{tab:gate_ablation}).
A one-way configuration that relies solely on cross-scale attention (CA) produces sharp local boundaries but exhibits limited texture continuity, often resulting in fragmented predictions in heterogeneous terrain regions.
Incorporating texture-based cues (TB) improves structural stability by reinforcing fine-scale appearance information; however, without explicit coordination with global semantics, overall consistency remains limited.

In contrast, the three-way configuration \(\{\mathrm{CA}, \mathrm{TB}, T_0\}\) achieves the most balanced interaction by explicitly coordinating cross-scale semantics, structural cues, and the original semantic token representation.
This configuration yields more coherent predictions across terrain transition areas and is therefore adopted as the default setting for all subsequent experiments.

\vspace{2pt}
\noindent\textbf{Robustness to Annotation Noise.}
We evaluate robustness to annotation noise by training models on synthetically perturbed RUGD labels and testing on clean annotations (Table~\ref{tab:noise_robustness_compare}), following common practice for studying learning under noisy supervision~\cite{ghosh2017robust, han2018co, pranto2022effect}.
This setting reflects realistic off-road scenarios where ground-truth boundaries are often imprecise or internally inconsistent due to vegetation clutter, reflections, and weak terrain transitions.

As shown in Fig.~\ref{fig:qual_gtnoise}, misaligned or noisy ground-truth labels cause GA-Nav to produce fragmented regions and unstable boundary patterns.
In contrast, CSTR maintains coherent region interiors and structurally consistent predictions even when supervision is unreliable.
This qualitative behavior indicates that consolidating global semantics before applying selective structural correction prevents the propagation of label noise into the final prediction.

The quantitative results in Table~\ref{tab:noise_robustness_compare} further support this observation.
As the noise radius increases, CSTR exhibits consistently smaller performance degradation than GA-Nav across mIoU, bIoU, and \(F1_{\text{boundary}}\), demonstrating improved robustness under imperfect supervision.
Notably, the advantage of CSTR becomes more pronounced in boundary-sensitive metrics, suggesting that the proposed decoder effectively suppresses noise amplification near ambiguous class transitions.

Together, the qualitative evidence in Fig.~\ref{fig:qual_gtnoise} and the quantitative trends in Table~\ref{tab:noise_robustness_compare} confirm that the proposed design is inherently robust to annotation noise.
By stabilizing semantic representations at a compact bottleneck and selectively correcting fine-scale ambiguities through gated interaction and point-wise refinement, CSTR avoids overfitting to erroneous labels and yields reliable predictions in challenging off-road environments.

\section{Conclusion}

We presented a noise-robust decoding framework tailored for off-road semantic segmentation under thick boundaries, class imbalance, and ambiguous supervision.
The proposed Cross-Scale Decoder consolidates multi-scale features into a compact semantic core via Global--Local Token Refinement (GLTR), extracts boundary-relevant structural cues through Boundary-Guided Correction (BGC), and selectively integrates them using Gated Cross-Scale Interaction (GCS) with uncertainty-guided point-wise refinement.
A boundary-band regularizer further stabilizes fragile transition regions during training without introducing additional inference cost.

This design is effective because it concentrates computation on stable semantic representations while correcting structural ambiguities only where necessary, thereby avoiding noise amplification.
Single-shot gated interaction preserves boundary geometry without repeated dense fusion, and sparse point-wise refinement focuses computation on uncertain regions, yielding both sharper predictions and efficient inference.

While minor failure cases remain in highly reflective or textureless regions with limited structural cues, the modular and noise-aware design naturally extends to settings with imperfect supervision and domain shifts.
Moreover, the framework can be integrated into broader perception pipelines that jointly reason about complementary cues such as terrain type or traversability, enabling practical deployment in off-road and field robotics.



\bibliographystyle{unsrt}
\bibliography{references}

\end{document}